\documentclass[a4paper]{article}

\usepackage{INTERSPEECH2022}
\usepackage{todonotes,cite}
\usepackage{nidanfloat}
\usepackage{enumitem}
\usepackage{multirow}
\usepackage{algorithm}

\usepackage{algorithmic}

\title{Streaming parallel transducer beam search with fast-slow cascaded encoders}
\name{Jay Mahadeokar\textsuperscript{*}, Yangyang Shi\textsuperscript{*}\thanks{*Equal Contribution}, Ke Li, Duc Le, Jiedan Zhu, Vikas Chandra, Ozlem Kalinli, Michael L Seltzer}

\address{Meta AI, USA}
\email{jaym@fb.com, yyshi@fb.com}

\begin{document}

\maketitle
\begin{abstract}
Streaming ASR with strict latency constraints is required in many speech recognition applications. In order to achieve the required latency, streaming ASR models sacrifice accuracy compared to non-streaming ASR models due to lack of future input context. Previous research has shown that streaming and non-streaming ASR for RNN Transducers can be unified by cascading causal and non-causal encoders. This work improves upon this cascaded encoders framework by leveraging two \emph{streaming} non-causal encoders with variable input context sizes that can produce outputs at different audio intervals (e.g. fast and slow). 
We propose a novel parallel time-synchronous beam search algorithm for transducers that decodes from fast-slow encoders, where the slow encoder corrects the mistakes generated from the fast encoder. 
The proposed algorithm, achieves up to 20\% WER reduction with a slight increase in token emission delays on the public Librispeech dataset and in-house datasets.
We also explore techniques to reduce the computation by distributing processing between the fast and slow encoders. Lastly, we explore sharing the parameters in the fast encoder to reduce the memory footprint. This enables low latency processing on edge devices with low computation cost and a low memory footprint. 
\end{abstract}


\section{Introduction}
\vspace{-1mm}

A responsive user experience is critical for voice-based virtual assistant applications. The latency of speech recognition determines the perceived system responsiveness for both voice commands (time from speech to action) and dictation (feeling of "snappiness").
Low latency requires streaming ASR, where incoming speech is processed incrementally based on partial context (while non-streaming ASR, e.g.~sequence-to-sequence models, run only after observing the whole utterance).

In the family of End-to-End (E2E) ASR models~\cite{Graves2012,he2019rnnt,Rao2017,Graves2014,Chan2016,ds2,Miao2016}, where acoustic model, pronunciation, and language model are combined into a single neural network, the recurrent neural network transducer, or RNN-T~\cite{Graves2012,he2019rnnt,Rao2017}, intrinsically supports for streaming. In~\cite{jiahui_2021_fastemit,jay_2020_arrnnt}, it is proposed to improve the RNN-T's token emission latency by sequence-level emission regularization and alignment restrictions, respectively. 

In~\cite{tara_2019_two_pass}, a non-streaming E2E LAS model~\cite{Chan2016} is applied for second-pass rescoring to compensate for the accuracy loss from the RNN-T's limited context. However, \cite{chiu_2019_long_form} shows that LAS-type models suffer from accuracy loss for long-form speech utterances compared to a non-streaming encoder-based RNN-T. Inspired by ``universal ASR'' \cite{yu2020universal}, the idea of cascaded encoders (causal and non-causal) with RNN-Ts are introduced in \cite{cascadedNarayanan}, where a non-streaming encoder is trained directly on the output of the streaming encoder instead of
input acoustic features allowing the non-streaming decoder to use fewer layers instead of a fully non-streaming model. \cite{li2021better} builds on this work by using a two-pass beam search. The first pass uses only the causal encoder, while during the second pass, additional non-causal layers utilize both the left and the right context of the 1st-pass encoder outputs as the input to a shared RNN-T decoder.

In other related work, \cite{hu2020deliberation} proposes to attend to both acoustics and first-pass hypotheses (``deliberation network''). \cite{yyshi_dynamic_2021} proposes to apply a subset of an encoder for the beginning part of utterance and a full encoder for the remaining utterance. \cite{wang2021deliberation} improves the align-refine approach introduced in \cite{chi2020align}
by using a cascaded encoder that captures more audio context
before refinement and alignment augmentation, which enforces
learning label dependency.

However, the non-streaming encoder has non-trivial user-perceived latency and memory footprint increase for long-form speech applications (e.g., dictation and messaging). We propose to improve the cascaded encoder framework such that \emph{both} encoders are streaming and non-causal, where for both encoders, look-ahead context is used. The proposed framework is called fast-slow cascaded encoders. The fast encoder produces outputs more frequently, while the slow encoder takes as input multiple segments output by the fast encoder and produces results with more extensive delays. We propose using a novel streaming parallel beam search that leverages both the fast and the slow encoders with shared search space. The fast encoder beam search produces timely partial results from fast encoder outputs to improve token emission delays. Whenever the slow encoder outputs are available, the slow encoder beam search updates the partial output results, at the same time also updating the candidates considered by fast encoder beam search. 

Running parallel beam search with fast-slow encoders has real-time factor and memory implications. We carefully analyze these run-time constraints and propose techniques to improve them by distributing parameters across fast-slow encoders and using smaller beam sizes for fast encoders. Similar to \cite{dehghani2018universal, li2019improving,dabre2019recurrent,sho_layer_sharing_2021} we also explore sharing of parameters across layers to reduce the memory footprint.

\section{Methodology}
This section describes the model architecture, training, and decoding procedures for the proposed model. Similar to the cascaded encoder work~\cite{cascadedNarayanan}, the proposed method focuses on the encoder in the RNN-T framework~\cite{Graves2012}.

\subsection{Streaming Fast Slow Cascaded Encoders}
\begin{figure}[h]
\vspace{-2.5mm}
    \centering
    \includegraphics[width=0.3\textwidth]{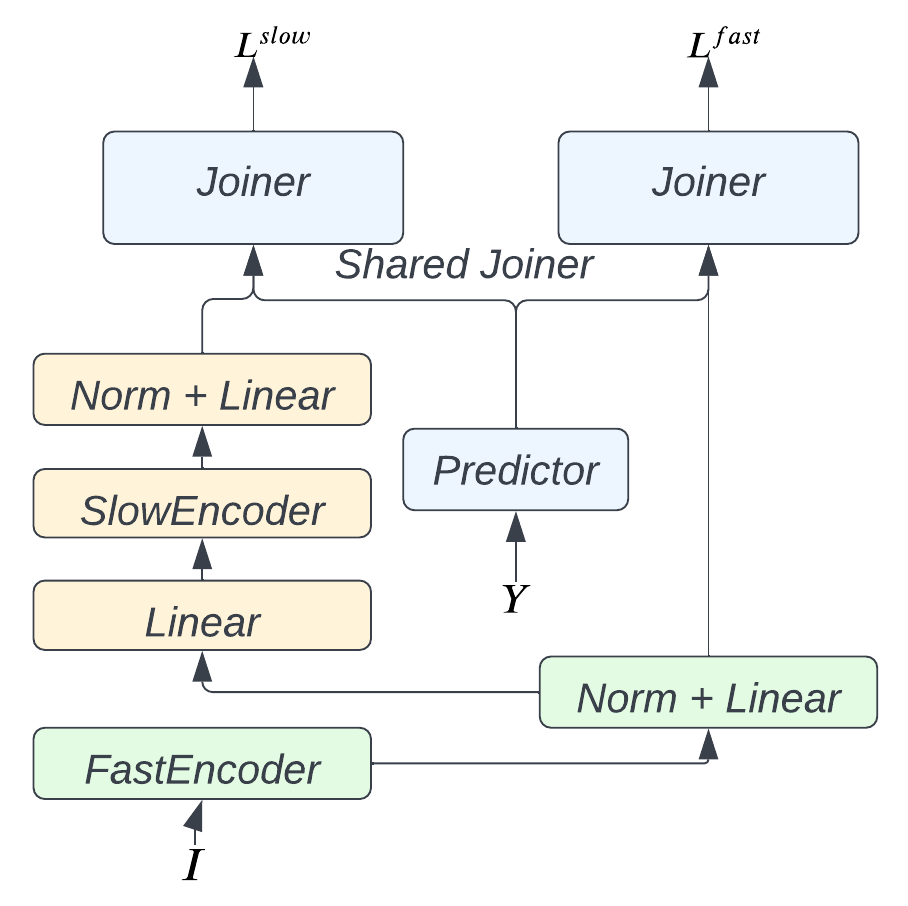}
    \caption{Illustration for streaming cascaded encoder using fast-slow encoders. The joiner and predictor are shared for both fast encoder and slow encoder.}
    \label{fig:illustration_cascaded_encoder}
    \vspace{-1.7em}
\end{figure}

Figure~\ref{fig:illustration_cascaded_encoder} gives the illustration of the streaming cascaded encoder. Different from the work~\cite{cascadedNarayanan} where the causal encoder and the non-causal encoder stochastically use different training samples within a minibatch, in this work, both encoders leverage the same training data. Rather than cascading a non-streaming encoder with a causal encoder in ~\cite{cascadedNarayanan}, both encoders in our framework are streaming non-causal encoders. 

As shown in Figure~\ref{fig:illustration_cascaded_encoder}, given the audio feature input $I$, the cascaded encoder generates two representations, one from the fast encoder and the other from the slow encoder. The same joiner and predictor in the RNN-T model take the representations and then give the logits for the audio and transcript pairs in training. The losses $L^{\mathrm{fast}}$ and $L^{\mathrm{slow}}$ are for the fast encoder and the slow encoder, respectively. In training, the final loss is 
\begin{align}
L &= L^{\mathrm{slow}} + \lambda L^{\mathrm{fast}}
\end{align}
where $0<\lambda < 1$. 
The weighted loss not only passes gradient back for both the fast and the slow encoder but also stabilizes the training, especially for deep encoder structure~\cite{Andros2019}.

Both the fast encoder and the slow encoder use a stack of Emformer~\cite{emformer} layers that apply the block processing method to support streaming ASR. The block processing method segments the whole input sequence $I$ into multiple blocks. Each block is padded with the corresponding right context (lookahead context). The Emformer stores the self-attention keys and values of the history context in the states to save computations. In Fig.~\ref{fig:illustration_cascaded_beam_Search}, the fast encoder uses segment size 4 and right context 1. The slow encoder uses the same right context size but double the segment size. Given the current block $C_j = [I^{\mathrm{fast}}_{4j},...,I^{\mathrm{fast}}_{4j+3}]$, the right context $R_j = I^{\mathrm{fast}}_{4j+4}$ and the history context state $H_j$, the fast encoder gives the following outputs for the current block and the right context.
\begin{align}
&[O^{\mathrm{fast}}_{4j},...,O^{\mathrm{fast}}_{4j+3}] = \mathtt{encodeFast}(C_j,[C_j, R_j, H_j]) \\
&\hat{O}^{\mathrm{fast}}_{{4j+4}} = \mathtt{encodeFast}(I^{\mathrm{fast}}_{{4j+4}},[C_j, R_j, H_j])
\end{align}
Note $\hat{O}^{\mathrm{fast}}_{{4j+4}}$ is the result from the attention of $I^{\mathrm{fast}}_{{4j+4}}$ with $j$-th segment context and $O^{\mathrm{fast}}_{{4j+4}}$ uses the attention of $I^{\mathrm{fast}}_{{4j+4}}$ with $(j+1)$-th segment context. Figure~\ref{fig:illustration_cascaded_beam_Search} shows that slow encoder takes the outputs $[O^{\mathrm{fast}}_{{4j}},...,O^{\mathrm{fast}}_{{4j+7}}]$ from twice the fast encoder forward processes as input and the right context output $\hat{O}^{\mathrm{fast}}_{{4j+8}}$.

\subsection{Parallel Beam Search}
For RNN Transducers, \cite{Graves2012} outlines a beam search algorithm. Let's assume that the algorithm uses a search space $\Gamma$. During beam search we iterate over audio time-steps $t = {0 \dots T}$ and search for $N$ (beam-size) most probable ASR hypothesis using sets $A$ and $B$. $A$ contains the current best hypothesis at time $t$, while the set $B$ stores the most probable candidates for step $t+1$. We reuse the beam search algorithm and extend it for fast-slow cascaded encoders as outlined in Algorithm \ref{parallel_beamsearch_algo}.

\begin{algorithm}[t] 
\caption{Parallel beam search for cascaded encoders.} 
\label{parallel_beamsearch_algo} 
\begin{algorithmic} 

\STATE $T^{\mathrm{fast}} =  \mathtt{fastSegmentSize}()$ 

\STATE $T^{\mathrm{slow}} = \mathtt{slowSegmentSize}()$
\STATE $N^{\mathrm{fast}} = \mathtt{fastBeamSize}()$
\STATE $N^{\mathrm{slow}} = \mathtt{slowBeamSize}()$
\STATE $B^{\mathrm{fast}} \gets \varnothing ; B^{\mathrm{slow}} \gets \varnothing$
\STATE $(H^{\mathrm{fast}}, H^{\mathrm{slow}}) \gets \mathtt{initModelState}()$
\STATE $\Gamma \gets \mathtt{initSearchSpace}()$
\STATE $I^{\mathrm{slow}} \gets \varnothing$
\FOR {$t = T^{\mathrm{fast}}$ to $T$ by $T^{\mathrm{fast}}$}
    \STATE $(I^{\mathrm{fast}}, R^{\mathrm{fast}}) = \mathtt{getFeatures}(t, t - T^{\mathrm{fast}})$ 
    \STATE $(O^{\mathrm{fast}}, H^{\mathrm{fast}}) = \mathtt{encodeFast}(I^{\mathrm{fast}}, (I^{\mathrm{fast}}, R^{\mathrm{fast}}, H^{\mathrm{fast}}))$
    \STATE $B^{\mathrm{fast}} \gets \mathtt{beamSearch}(O^{\mathrm{fast}}, B^{\mathrm{fast}}, N^{\mathrm{fast}}, \Gamma, H^{\mathrm{fast}})$
    \STATE $I^{\mathrm{slow}} \gets \mathtt{concat}(I^{\mathrm{slow}}, O^{\mathrm{fast}})$
    \IF{ $t \mod T^{\mathrm{slow}} = 0$ or $t = T$}
        \STATE $(O^{\mathrm{slow}}, H^{\mathrm{slow}}) = \mathtt{encodeSlow}(I^{\mathrm{slow}}, (I^{\mathrm{slow}}, R^{\mathrm{slow}}, H^{\mathrm{slow}}))$
        \STATE $B^{\mathrm{slow}} \gets \mathtt{beamSearch}(O^{\mathrm{slow}}, B^{\mathrm{slow}}, N^{\mathrm{slow}}, \Gamma, H^{\mathrm{slow}})$
        \STATE $I^{\mathrm{slow}} \gets \varnothing$
        \STATE $B^{\mathrm{fast}} \gets B^{\mathrm{slow}}$
    \ENDIF
\ENDFOR
\RETURN  $y$ with highest $\log Pr(y) / |y| $ in $B^{\mathrm{slow}}$ 
\end{algorithmic}
\end{algorithm}

\begin{figure}[b]
    \includegraphics[width=0.45\textwidth]{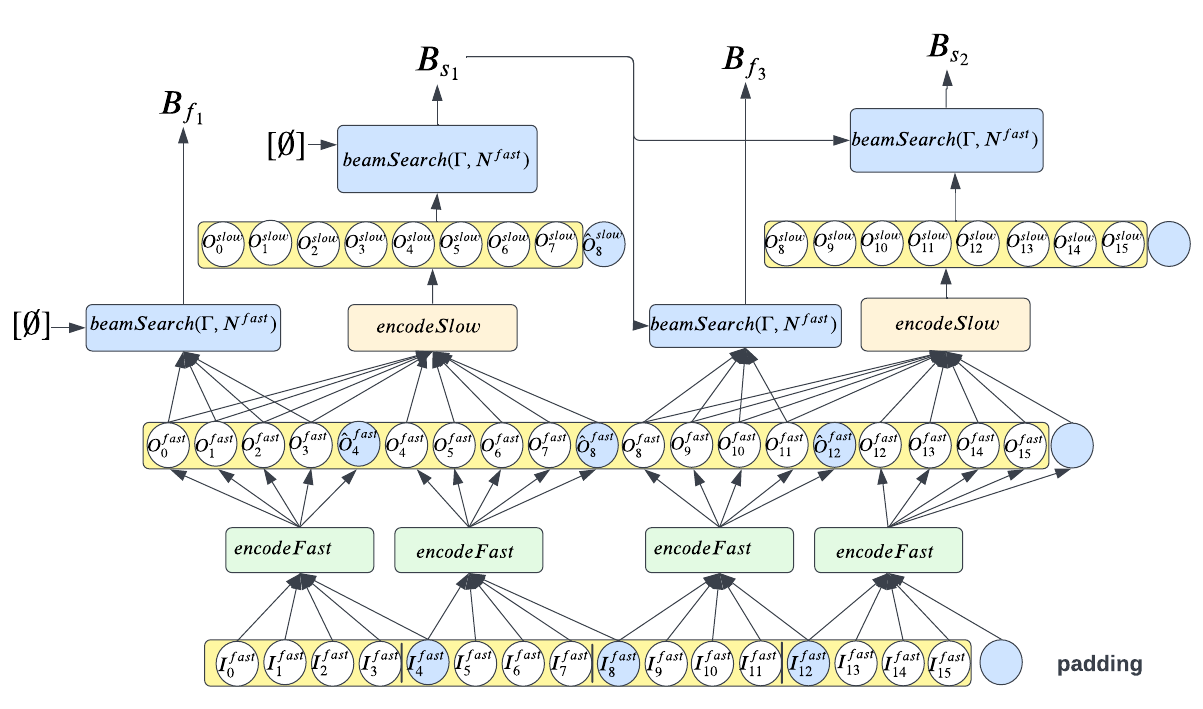}
    \caption{Illustration for parallel beam search with cascaded encoders. Both the fast encoder and the slow encoder uses the same right context size 1. The fast encoder and the slow encoder use segment size 4 and 8, respectively.}
    \label{fig:illustration_cascaded_beam_Search}
    \vspace{-1.7em}
\end{figure}

We maintain two sets, $B^{\mathrm{fast}}$ and $B^{\mathrm{slow}}$, that represent the best hypotheses generated using fast-slow encoder outputs, respectively. Both the fast and the slow encoders use Emformer layers which use states to store the key and value for the history left context. Let $H^{\mathrm{fast}}$ and $H^{\mathrm{slow}}$ denote these states. Let $T^{\mathrm{fast}}$, $T^{\mathrm{slow}}$ denote the segment size and $N^{\mathrm{fast}}$, $N^{\mathrm{slow}}$ denote the beam size used for fast-slow encoder beam search. We iterate over the audio time-steps in the interval of $T^{\mathrm{fast}}$ and run a fast encoder. The fast encoder outputs $O^{\mathrm{fast}}$ and updates the state $H^{\mathrm{fast}}$. $O^{\mathrm{fast}}$ is used to call beam search to update $B^{\mathrm{fast}}$ which contains the current partial hypothesis output by ASR. At the same time, $O^{\mathrm{fast}}$ is concatenated with previously cached vector $I^{\mathrm{slow}}$, which forms the input to slow encoder. We skip details of the right context $\hat{O^{\mathrm{fast}}}$ in Algorithm \ref{parallel_beamsearch_algo} for simplicity.
  
When we have processed $T^{\mathrm{slow}}$ time-steps, the slow encoder is called with $I^{\mathrm{slow}}$ to produce slow encoder output $O^{\mathrm{slow}}$ and state $H^{\mathrm{slow}}$, which is then used to update $B^{\mathrm{slow}}$ using beam search, that shares the search space $\Gamma$. Shared search space $\Gamma$ is crucial for efficient run-time implementation. We then update the set $B^{\mathrm{fast}}$ with $B^{\mathrm{slow}}$ which typically corrects the outputs from $B^{\mathrm{fast}}$, and discard existing $B^{\mathrm{fast}}$ hypothesis. In the end, we return $y$, which is the most probable hypothesis in $B^{\mathrm{slow}}$. Figure \ref{fig:illustration_cascaded_beam_Search} illustrates this using example of 4 fast encoder calls and 2 slow encoder calls.

\section{Experimental Setup}
\label{sec:experimental_setup}
\vspace{-2mm}

\subsection{Datasets}

\subsubsection{Librispeech}
\label{subsec:librispeech}
The Librispeech~\cite{panayotov2015librispeech} corpus contains 970 hours of labeled speech. We extract 80-channel filterbanks features computed from a 25 ms window with a stride of 10 ms. We apply
spectrum augmentation (SpecAugment ~\cite{park2019specaugment}) with mask parameter $F$ = 27, ten time masks with maximum time-mask ratio $p_S$ = 0.05, and speed perturbation. 

\subsubsection{Large-Scale In-House Data}
\label{subsec:in_house_data}
Our in-house training set combines two sources. The first consists of 20K hours of English video data publicly shared by Facebook users; all videos are completely de-identified before transcription. The second contains 20K hours of manually transcribed de-identified English data with no user-identifiable information (UII) in the voice assistant domain. All utterances are morphed when researchers manually access them to further de-identify the speaker. Note that the data are not morphed during training. We further augment the data with speed perturbation, simulated room impulse response, and background noise, resulting in 83M utterances (145K hours).

We consider three in-house evaluation sets:

\textbf{VA1} -- 10.2K hand-transcribed de-identified short-form utterances (less than five words on average) in the voice assistant domain, collected from internal volunteer participants. The participants consist of households that have consented to have their Portal voice activity reviewed and analyzed.

\textbf{VA2} -- 44.2K hand-transcribed de-identified short-form utterances in the voice assistant domain, collected by a third-party data vendor via Oculus devices.

\textbf{Q\&A} -- 5.7K hand-transcribed de-identified medium-length utterances (more than 13 words on average) collected by crowd-sourced workers via mobile devices. The utterances consist of free-form questions directed toward a voice assistant.

\subsection{Evaluation Metrics}\label{evalmetrics}
To measure the model's performance and analyze trade-offs, we track the following metrics: 

\textbf{Accuracy:} We use word-error-rate (WER) to measure model accuracy on evaluation sets.  

\textbf{Emission Delay:} (or finalization delay) as defined in \cite{jay_2020_arrnnt} is the audio duration between the time when the user finished speaking the ASR token, and the time when the ASR token was surfaced as part of the 1-best partial hypothesis, also referred to as emission latency in~\cite{yu2021dualmode}. We track the Average ($\mathrm{ED}_{Avg}$) and P99 ($\mathrm{ED}_{P99}$) token emission Delays. 

\textbf{Correction rate:} Our proposed technique uses a slow encoder to correct mistakes made by the fast encoder.  Let $\mathrm{WER}^{\mathrm{fast}}$ be the word error rate if we use fast encoder's output and $\mathrm{WER}^{\mathrm{slow}}$ be the word error rate when using slow encoder's output. We define correction rate (CR) as CR = $\mathrm{WER}^{\mathrm{fast}}$ - $\mathrm{WER}^{\mathrm{slow}}$.

\textbf{Real Time Factor:} To measure the impact of parallel beam search on run-time / compute, we use Real Time Factor (RTF) measured on an actual android device.

\subsection{Model setup}
We use an RNN-T model architecture that has emformer \cite{shi2021emformer} as encoders. We use a stacked time reduction layer with a stride of 4, which converts 80-dimensional input features into 320-dimensional features that are input to the encoder. Predictor consists of 3 LSTM layers, with Layer Norm having 512 hidden units. Both encoder and predictor project embeddings of 1024 dimensions, which are input to joint layer, consist of a simple DNN layer and a softmax layer, predicting a word-piece output of size 5k dimensions. 

For librispeech experiments, we train our models for 120 epochs. We use an ADAM optimizer and a tri-stage LR scheduler with a base learning rate of 0.001, a warmup of 10K iterations, and forced annealing after 60K epochs. Experiments on in-house data follow a similar model architecture and training hyperparameters. Models are trained for 15 epochs on large-scale training data. 

\section{Results}
\subsection{Optimizing WER and latency}
\subsubsection{Effects of varying slow encoder context}
In table \ref{table:librispeech_result_encoders} we train baselines B1 to B5 with 20 layer models with varying context size of 160 to 6400 ms. As expected, with increased model context, we see improved WERs and increased emission delays. We train models using streaming cascaded encoders (C1 to C4) with 15 fast layers and a fixed context of 160ms while the five slow layers are trained with varying contexts. Using streaming parallel beam search, we achieve up to 20\% WERR (B1 Vs. C4). As shown by CR metric, since we correct ~2.63\% words with C4, the ED P99 degrades from 560 to 800ms, with minimal effect on ED Avg. 

\def\ccbf#1{\multicolumn{1}{c|}{\textbf{#1}}}
\def\ccbff#1{\multicolumn{1}{c}{\textbf{#1}}}
\def\ccbfb#1{\multicolumn{1}{|c|}{\textbf{#1}}}
\def\d{\hphantom{0}}

\begin{table}[h]
\centering
\begin{tabular}{|p{0.09cm}| p{0.6cm} |p{0.85cm}|p{0.6cm}  p{0.6cm} | p{0.45cm} p{0.5cm} | p{0.45cm} |}\hline
    &\ccbf{Model} & \ccbf{Con-} &\ccbff{Test-} & \ccbf{Test-} & \multicolumn{2}{c}{ED} & \ccbfb{CR} \\
    & \ccbf{}   & \ccbf{text} &\ccbff{clean} & \ccbf{other} &  Avg & P99 &  \\\hline\hline

    \scalebox{0.7}[0.7]{B1} &  &\scalebox{.8}[.8]{{\d}160} & 3.46 & 8.96 & {\d}335 & {\d}560 &\\ 

     \scalebox{0.7}[0.7]{B2} & \scalebox{.8}[.8]{20 full} &\scalebox{.8}[0.8]{{\d}800} & 3.15 & 8.10 & {\d}651 & 1160 & \\
     \scalebox{0.7}[0.7]{B3} & &\scalebox{0.8}[0.8]{1600} & 3.17 & 7.63 & {\d}971 & 1880 & N/A \\ 

      \scalebox{0.7}[0.7]{B4} & &\scalebox{0.8}[0.8]{3200} & 3.11 & 7.23 & 1612 & 3360 & \\ 

       \scalebox{0.7}[0.7]{B5} & &\scalebox{0.8}[0.8]{6400} & 3.10 & 6.99 & 2754 & 6276 & \\

     \hline\hline
     \scalebox{0.7}[0.7]{C1} & &\scalebox{0.8}[0.8]{160/{\d}800} & 3.22 & 8.24 & {\d}336 & {\d}600 & 1.38 \\
     \scalebox{0.7}[0.7]{C2} &\scalebox{.8}[.8]{15 fast} &\scalebox{0.8}[0.8]{160/1600} & 3.11 & 7.83 & {\d}329 & {\d}600 & 1.74\\ 
     \scalebox{0.7}[0.7]{C3} &\scalebox{.8}[.8]{5 slow } &\scalebox{0.8}[0.8]{160/3200} & 2.99 & 7.28 & {\d}329 & {\d}600 & 2.39 \\ 
     \scalebox{0.7}[0.7]{C4} & &\scalebox{0.8}[0.8]{160/6400} & 2.91 & 7.15 & {\d}346 & {\d}800 & 2.63 \\ \hline

\end{tabular}
\vspace{0.5em}
\caption{Experiments comparing baseline models trained using different context sizes and fixed fast encoder context of 160ms, with varying slow-encoder context. CR and ED are using Test-other.}
\label{table:librispeech_result_encoders}
\vspace{-2.0em}
\end{table}

\subsubsection{Further improving latency}
In this section, we explore further improving the token emission latency of the model. \cite{jay_2020_arrnnt} shows that emission latency can be controlled by restricting optimized paths while also reducing compute and improving training throughput. Fast-emit \cite{yu2021fastemit} introduces regularization to force timely token emissions. We empirically verify that applying fast-emit regularization on a restricted set of paths gives the best of both worlds in terms of faster latency and optimal throughput. All experiments in Table \ref{table:librispeech_result_latency} use 15 fast and 5 slow encoder layers and AR-RNNT left and right restrictions of 0ms and 600ms. Using larger fast-emit $\lambda$, we reduce ED Avg from 329 to 174ms with some degradation on test-other WER. We also explore reducing the context of a slow encoder to improve token emission delay further, as shown in experiments L4 and L5.

\begin{table}[h]
\centering
\begin{tabular}{|p{0.1cm}| p{0.8cm} |p{1.0cm}|p{0.8cm}p{0.8cm}| p{0.8cm} p{0.8cm} |}\hline
    &\ccbf{Con-} & \ccbf{Fast-} &\ccbff{Test-} & \ccbf{Test-} & \multicolumn{2}{c|}{ED} \\
    &  \ccbf{text}  & \ccbf{emit $\lambda$} &\ccbff{clean} & \ccbf{other} &  Avg & P99   \\\hline\hline

    \scalebox{0.7}[0.7]{L1} & 160 / & 0.0 & 2.99 & 7.28 & 329 & 600\\ 

     \scalebox{0.7}[0.7]{L2} & 3200 & 0.001 & 3.02 & 7.47 & 299 & 600 \\ 


       \scalebox{0.7}[0.7]{L3} & & 0.01 & 3.07 & 7.56 & 174 & 600 \\

     \hline\hline
     \scalebox{0.7}[0.7]{L4} & 80 / & 0.0 & 3.01 & 7.6 & 295 & 520\\
     \scalebox{0.7}[0.7]{L5} & 3200 & 0.01 & 3.09 & 7.72 & 135 & 560 \\ 
     \hline
     
\end{tabular}
\vspace{0.5em}
\caption{Experiments with varying fast-emit lambda and smaller fast encoder context. ED is computed using test-other.}
\label{table:librispeech_result_latency}
\vspace{-1.0em}
\end{table}
\vspace{-1.0em}
\subsection{Optimizing Runtime}
\subsubsection{Distributing layers between fast / slow encoder}
The parallel beam search using the fast-slow encoders impacts the model's real-time factor (RTF). Experiments in Table \ref{table:rtf_improvements} look into techniques to improve the RTF of the proposed model by analyzing the effects of distributing layers between fast-slow encoders and reducing the beam size of fast encoder search. Models R1 to R4 are trained using 160ms and 800ms context sizes for fast-slow encoders. We observe that since a slow encoder has a larger context compared to the fast encoder, it incurs less compute due to overlapping right context, and the execution can be batched across timesteps more efficiently. Combining this with a reduced beam size of fast encoder beam search, we see improvements to RTF (0.55 to 0.48 for B1 to R3). Configuration R3 provides the best tradeoffs in terms of WER (8.13), and P99 ED (600ms), RTF (0.48). Note that similar to Table \ref{table:librispeech_result_encoders} Avg ED is mostly unchanged for R1 to R4 compared to B1. Further optimization of runtime implementation and other techniques like applying additional time reduction layer within slow encoder can further improve RTF, which we plan to explore as future work.

\begin{table}[h]
\centering
\begin{tabular}{|p{0.1cm}| p{0.6cm} |p{0.5cm}| p{0.5cm} p{0.5cm}  p{0.5cm}| p{0.5cm} p{0.5cm}p{0.5cm} |}\hline
    &\textbf{\scalebox{0.8}[0.8]{Layers}} &Test  &Beam &  10 & ED &Beam &  2 & ED\\
    & & other &  RTF & CR & P99 & RTF & CR & P99 \\\hline\hline
    \scalebox{0.7}[0.7]{B1} & 20 & 8.96 & 0.44 &  & 560 & &  & \\ 
    \hline \hline

     \scalebox{0.7}[0.7]{R1} & 3+17 & 8.2  & 0.55 & 15.6 & 1120 & 0.42 & 23.3 & 1120 \\

       \scalebox{0.7}[0.7]{R2} & 7+13 & 8.54  & 0.55 & 4.6 & 880 &   0.45 & 8.49 & 960\\

     \scalebox{0.7}[0.7]{R3} & 13+7 & 8.13 & 0.56 & 2.16  & 600  & 0.48 & 4.84 & 600 \\
    \scalebox{0.7}[0.7]{R4
    } & 17+3 & 8.45 & 0.55 & 0.81 & 560 & 0.5 & 2.42 & 600\\
     \hline
     
\end{tabular}
\vspace{0.5em}
\caption{Experiments to analyze WER Vs. RTF tradeoffs by distributing the layers between the fast and the slow encoders and using smaller beam for fast encoder outputs.}
\label{table:rtf_improvements}
\vspace{-1.5em}
\end{table}

\subsubsection{Sharing parameters for memory reduction}
Memory optimizations are critical for on-the-edge applications. We explore sharing the parameters of layers in fast encoders to further improve the memory consumption similar to \cite{li2019improving}. Our intuition is that since the slow encoder has access to extra future constraints, the same parameters could be utilized to further correct the mistakes made by fast encoders. 

We explored different layer sharing in fast-slow cascaded frameworks, such as sharing layers between the fast and the slow encoders, sharing layers in slow encoders, and sharing layers in fast encoders. We found that sharing layers in a fast encoder gave better performance. In Table~\ref{table:librispeech_result_share_memory}, we only list the results from layer sharing in a fast encoder situation, specifically, sharing continuous 13 layers from the 2nd layer to the 14th layer. 

In Table~\ref{table:librispeech_result_share_memory}, using layer sharing on top of baseline models (P1) shows significant WER reduction compared with the same model size baseline (B2). Similar to the trend in Table~\ref{table:librispeech_result_encoders}, leveraging 800 ms context in the slow encoder (Q1) gets more than 5$\%$ relative WER reduction over layer sharing on baseline models (P1). Further Extending the slow encoder with context to 6.4 s on top of the layer sharing, the 41 million parameters model even outperforms the 79 million parameters model by relative WER reduction 13$\%$ on test-other and 9$\%$ on test-clean.

\begin{table}[h]
\centering
\begin{tabular}{|p{0.08cm}| p{0.6cm} |p{0.85cm}|p{0.85cm}|p{0.85cm}|p{0.6cm} p{0.6cm}| }\hline
    &\textbf{\scalebox{0.8}[0.8]{Model}} & \textbf{\scalebox{0.8}[0.8]{Context }} & \textbf{\scalebox{0.8}[0.8]{\#params}}& \textbf{layers-share}&\textbf{Test-clean} & \textbf{Test-other} \\\hline\hline
    \scalebox{0.7}[0.7]{B1} & \scalebox{.8}[.8]{20 full} &\scalebox{.8}[.8]{160}&79M & - & 3.46 & 8.96 \\
    \scalebox{0.7}[0.7]{B2} & \scalebox{.8}[.8]{8 full} &\scalebox{.8}[.8]{160}&41M &- &4.32  & 11.05  \\ \hline \hline
    \scalebox{0.7}[0.7]{P1} & \scalebox{.8}[.8]{20} &\scalebox{.8}[.8]{160}&41M &2-14 &3.82  &9.50  \\ \hline \hline
    \scalebox{0.7}[0.7]{Q1} & \scalebox{.8}[.8]{15 fast} &\scalebox{.8}[.8]{160/800}&41M &2-14 & 3.55 & 9.04 \\
    \scalebox{0.7}[0.7]{Q2} &  \scalebox{.8}[.8]{5 slow}  &\scalebox{.8}[.8]{160/6400}&41M &2-14 & 3.16 & 7.86  \\\hline
\end{tabular}
\vspace{0.5em}
\caption{Experiments comparing baseline models trained using different number of parameters, layer sharing and layer sharing in fast slow cascaded encoder.}
\label{table:librispeech_result_share_memory}
\vspace{-1.5em}
\end{table}

\subsection{In-house dataset}
This section runs the most promising configurations on the in-house dataset. Experiment in Table \ref{table:in_house_expts} trains a baseline model using 20 layers. Similar to Table \ref{table:librispeech_result_encoders}, experiments P2 to P5 outline results using the proposed technique with 15 fast and 5 slow layers while varying slow encoder context. We see significant gains on VA2 and Q\&A domains, consisting of longer-form utterances than the VA1 dataset. We see 14.7\% and 10.7\% WERR on Q\&A and VA2 datasets comparing P1 Vs. P5, with minimal change to Avg ED, or P99 ED (not shown in table).  

\begin{table}[h]
\centering
\begin{tabular}{|p{0.08cm}| p{0.6cm} |p{0.85cm}|p{0.6cm} p{0.6cm} p{0.6cm} | p{0.6cm}  p{0.45cm} |}\hline
    &\textbf{\scalebox{0.8}[0.8]{Model}} & \textbf{\scalebox{0.8}[0.8]{Context }} &\textbf{VA1} & \textbf{Q\&A} &  \textbf{VA2} & ED Avg & CR \\\hline
    \scalebox{0.7}[0.7]{P1} & \scalebox{.8}[.8]{20 full} &\scalebox{.8}[.8]{160} & 4.71  & 7.51 & 13.35 & 388 &\\ 

     \hline
     \scalebox{0.7}[0.7]{P2} & &\scalebox{0.8}[0.8]{160/800} & 4.65 & 7.16 & 12.72 & 372 & 1.5  \\
     
     \scalebox{0.7}[0.7]{P3} & \scalebox{.8}[.8]{15 fast }&\scalebox{0.8}[0.8]{160/1600} &   4.57 & 6.82 & 12.13 & 371 & 1.88\\
     \scalebox{0.7}[0.7]{P4} &\scalebox{.8}[.8]{5 slow } &\scalebox{0.8}[0.8]{160/3200} & 4.66 & 6.66 & 12.05 &  381&  1.95\\ 
     
     \scalebox{0.7}[0.7]{P5} & &\scalebox{0.8}[0.8]{160/6400} & 4.72  & 6.4 & 11.92 & 410 &  2.29\\
   \hline

\end{tabular}
\vspace{0.5em}
\caption{Experiments on in-house dataset with different context size for slow encoder. ED and CR are computed on Q\&A dataset.}
\label{table:in_house_expts}
\vspace{-1.0em}
\end{table}

\vspace{-2.0em}
\section{Conclusion}
We proposed a framework that uses streaming parallel transducer beam search with fast-slow cascaded encoders. We show that using the proposed technique, achieving 15 to 20\% WER reduction on librispeech and in-house datasets, with trivial degradation to average token emission delays. We empirically show that additional techniques can improve model's runtime and memory. In future work, we will further explore optimizing the runtime by subsampling in the slow encoder. 

\textbf{Acknowledgement: } We would like to thank Frank Seide for careful review and feedback on the paper.

\bibliographystyle{IEEEtran}

\bibliography{template}

\begin{thebibliography}{10}
\providecommand{\url}[1]{#1}
\csname url@samestyle\endcsname
\providecommand{\newblock}{\relax}
\providecommand{\bibinfo}[2]{#2}
\providecommand{\BIBentrySTDinterwordspacing}{\spaceskip=0pt\relax}
\providecommand{\BIBentryALTinterwordstretchfactor}{4}
\providecommand{\BIBentryALTinterwordspacing}{\spaceskip=\fontdimen2\font plus
\BIBentryALTinterwordstretchfactor\fontdimen3\font minus
  \fontdimen4\font\relax}
\providecommand{\BIBforeignlanguage}[2]{{%
\expandafter\ifx\csname l@#1\endcsname\relax
\typeout{** WARNING: IEEEtran.bst: No hyphenation pattern has been}%
\typeout{** loaded for the language `#1'. Using the pattern for}%
\typeout{** the default language instead.}%
\else
\language=\csname l@#1\endcsname
\fi
#2}}
\providecommand{\BIBdecl}{\relax}
\BIBdecl

\bibitem{Graves2012}
A.~Graves, ``{Sequence Transduction with Recurrent Neural Networks},''
  \emph{arXiv preprint arXiv:1211.3711}, 2012.

\bibitem{he2019rnnt}
Y.~He, T.~N. Sainath, R.~Prabhavalkar, and Others, ``{Streaming End-to-end
  Speech Recognition for Mobile Devices},'' in \emph{Proc. ICASSP}, 2019.

\bibitem{Rao2017}
K.~Rao, H.~Sak, and R.~Prabhavalkar, ``{Exploring architectures, data and units
  for streaming end-to-end speech recognition with RNN-transducer},'' in
  \emph{Proc. ASRU}, 2018.

\bibitem{Graves2014}
A.~Graves and N.~Jaitly, ``{Towards End-To-End Speech Recognition with
  Recurrent Neural Networks},'' in \emph{Proc. JMLR}, 2014.

\bibitem{Chan2016}
W.~Chan, N.~Jaitly, Q.~Le, and O.~Vinyals, ``{Listen, attend and spell: A
  neural network for large vocabulary conversational speech recognition},'' in
  \emph{Proc. ICASSP}, 2016.

\bibitem{ds2}
D.~Amodei, R.~Anubhai, E.~Battenberg, C.~Case, and Others, ``{Deep Speech 2:
  End-to-End Speech Recognition in English and Mandarin},'' \emph{arXiv
  preprint arXiv:1512.02595}, 2015.

\bibitem{Miao2016}
Y.~Miao, M.~Gowayyed, and F.~Metze, ``{EESEN: End-to-end speech recognition
  using deep RNN models and WFST-based decoding},'' in \emph{Proc. ASRU}, 2016.

\bibitem{jiahui_2021_fastemit}
J.~Yu, C.-C. Chiu, B.~Li, and Others, ``{Fastemit: Low-Latency Streaming Asr
  With Sequence-Level Emission Regularization},'' in \emph{Proc. ICASSP},
  vol.~53, no.~9, 2021.

\bibitem{jay_2020_arrnnt}
J.~Mahadeokar, Y.~Shangguan, D.~Le, and Others, ``{Alignment restricted
  streaming recurrent neural network transducer},'' in \emph{Proc. SLT}, 2021.

\bibitem{tara_2019_two_pass}
T.~N. Sainath, R.~Pang, D.~Rybach, Y.~He, R.~Prabhavalkar, W.~Li, M.~Visontai,
  Q.~Liang, T.~Strohman, Y.~Wu, I.~McGraw, and C.~C. Chiu, ``{Two-pass
  end-to-end speech recognition},'' \emph{Proc. Interspeech}, 2019.

\bibitem{chiu_2019_long_form}
\BIBentryALTinterwordspacing
C.-C. Chiu, W.~Han, Y.~Zhang, R.~Pang, S.~Kishchenko, P.~Nguyen, A.~Narayanan,
  H.~Liao, S.~Zhang, A.~Kannan, R.~Prabhavalkar, Z.~Chen, T.~Sainath, and
  Y.~Wu, ``{A comparison of end-to-end models for long-form speech
  recognition},'' \emph{Proc. ASRU}, 2019. [Online]. Available:
  \url{https://arxiv.org/abs/1911.02242v1}
\BIBentrySTDinterwordspacing

\bibitem{yu2020universal}
J.~Yu, W.~Han, A.~Gulati, C.-C. Chiu, B.~Li, T.~N. Sainath, Y.~Wu, and R.~Pang,
  ``Universal asr: Unify and improve streaming asr with full-context
  modeling,'' \emph{arXiv preprint arXiv:2010.06030}, 2020.

\bibitem{cascadedNarayanan}
A.~Narayanan, T.~N. Sainath, R.~Pang, J.~Yu, C.-C. Chiu, R.~Prabhavalkar,
  E.~Variani, and T.~Strohman, ``Cascaded encoders for unifying streaming and
  non-streaming asr,'' in \emph{Proc. ICASSP}, 2021, pp. 5629--5633.

\bibitem{li2021better}
B.~Li, A.~Gulati, J.~Yu, T.~N. Sainath, C.-C. Chiu, A.~Narayanan, S.-Y. Chang,
  R.~Pang, Y.~He, J.~Qin \emph{et~al.}, ``A better and faster end-to-end model
  for streaming asr,'' in \emph{Proc. ICASSP}, 2021.

\bibitem{hu2020deliberation}
K.~Hu, T.~N. Sainath, R.~Pang, and R.~Prabhavalkar, ``Deliberation model based
  two-pass end-to-end speech recognition,'' in \emph{Proc. ICASSP}, 2020.

\bibitem{yyshi_dynamic_2021}
Y.~Shi, V.~Nagaraja, C.~Wu, J.~Mahadeokar, D.~Le, R.~Prabhavalkar, A.~Xiao,
  C.~F. Yeh, J.~Chan, C.~Fuegen, O.~Kalinli, and M.~L. Seltzer, ``{Dynamic
  encoder transducer: A flexible solution for trading off accuracy for
  latency},'' \emph{Proc. Interspeech}, 2021.

\bibitem{wang2021deliberation}
W.~Wang, K.~Hu, and T.~Sainath, ``Deliberation of streaming rnn-transducer by
  non-autoregressive decoding,'' \emph{arXiv preprint arXiv:2112.11442}, 2021.

\bibitem{chi2020align}
E.~A. Chi, J.~Salazar, and K.~Kirchhoff, ``Align-refine: Non-autoregressive
  speech recognition via iterative realignment,'' \emph{arXiv preprint
  arXiv:2010.14233}, 2020.

\bibitem{dehghani2018universal}
M.~Dehghani, S.~Gouws, O.~Vinyals, J.~Uszkoreit, and {\L}.~Kaiser, ``Universal
  transformers,'' \emph{arXiv preprint arXiv:1807.03819}, 2018.

\bibitem{li2019improving}
S.~Li, D.~Raj, X.~Lu, P.~Shen, T.~Kawahara, and H.~Kawai, ``Improving
  transformer-based speech recognition systems with compressed structure and
  speech attributes augmentation.'' in \emph{Proc. Interspeech}, 2019.

\bibitem{dabre2019recurrent}
R.~Dabre and A.~Fujita, ``Recurrent stacking of layers for compact neural
  machine translation models,'' in \emph{Proc. AAAI}, vol.~33, no.~01, 2019,
  pp. 6292--6299.

\bibitem{sho_layer_sharing_2021}
S.~Takase and S.~Kiyono, ``{Lessons on Parameter Sharing across Layers in
  Transformers},'' \emph{arXiv preprint arXiv 2104.06022}, 2021.

\bibitem{Andros2019}
A.~Tjandra, C.~Liu, F.~Zhang, and Others, ``{Deja-vu: Double Feature
  Presentation and Iterated loss in Deep Transformer Networks},'' \emph{Proc.
  ICASSP}, 2020.

\bibitem{emformer}
Y.~Shi, Y.~Wang, C.~Wu, C.-F. Yeh, and Others, ``{Emformer: Efficient Memory
  Transformer Based Acoustic Model For Low Latency Streaming Speech
  Recognition},'' in \emph{Proc. ICASSP}, 2021.

\bibitem{panayotov2015librispeech}
V.~Panayotov, G.~Chen, D.~Povey, and S.~Khudanpur, ``Librispeech: an asr corpus
  based on public domain audio books,'' in \emph{Proc. ICASSP}, 2015.

\bibitem{park2019specaugment}
D.~S. Park, W.~Chan, Y.~Zhang, C.-C. Chiu, B.~Zoph, E.~D. Cubuk, and Q.~V. Le,
  ``Specaugment: A simple data augmentation method for automatic speech
  recognition,'' \emph{arXiv preprint arXiv:1904.08779}, 2019.

\bibitem{yu2021dualmode}
J.~Yu, W.~Han, A.~Gulati, C.-C. Chiu, B.~Li, T.~N. Sainath, Y.~Wu, and R.~Pang,
  ``Dual-mode asr: Unify and improve streaming asr with full-context
  modeling,'' 2021.

\bibitem{shi2021emformer}
Y.~Shi, Y.~Wang, C.~Wu, C.-F. Yeh, J.~Chan, F.~Zhang, D.~Le, and M.~Seltzer,
  ``Emformer: Efficient memory transformer based acoustic model for low latency
  streaming speech recognition,'' in \emph{Proc. ICASSP}, 2021.

\bibitem{yu2021fastemit}
J.~Yu, C.-C. Chiu, B.~Li, S.-y. Chang, T.~N. Sainath, Y.~He, A.~Narayanan,
  W.~Han, A.~Gulati, Y.~Wu \emph{et~al.}, ``Fastemit: Low-latency streaming asr
  with sequence-level emission regularization,'' in \emph{Proc. ICASSP}, 2021.

\end{thebibliography}

\end{document}